%% file: main.tex
\documentclass[sigconf]{acmart}
\usepackage{multirow}
\usepackage{subcaption}
\usepackage{subcaption}
\usepackage{soul}
\usepackage[capitalize,nameinlink,noabbrev]{cleveref} % to emulate \autoref style

\AtBeginDocument{%
  }

\setcopyright{acmlicensed}
\copyrightyear{2025}
\acmYear{2025}
\acmDOI{XXXXXXX.XXXXXXX}
\acmConference[CIKM '25]{}{November 10--14, 2025}{Coex, Seoul, Korea}
\begin{document}

\title{AI on the Pulse: Real-Time Health Anomaly Detection with Wearable and Ambient Intelligence}

\author{Davide Gabrielli}
\affiliation{%
  \institution{Sapienza University of Rome}
  \city{Rome}
  \country{Italy}}
\email{davide.gabrielli@uniroma1.it}

\author{Bardh Prenkaj}
\affiliation{%
  \institution{Sapienza University of Rome}
  \city{Rome}
  \country{Italy}
}
\email{prenkaj@di.uniroma1.it}

\author{Paola Velardi}
\affiliation{%
  \institution{Sapienza University of Rome}
  \city{Rome}
  \country{Italy}}
\email{velardi@di.uniroma1.it}

\author{Stefano Faralli}
\affiliation{%
  \institution{Sapienza University of Rome}
  \city{Rome}
  \country{Italy}}
\email{faralli@di.uniroma1.it}

\begin{abstract}
We introduce AI on the Pulse, a real-world-ready anomaly detection system that continuously monitors patients using a fusion of wearable sensors, ambient intelligence, and advanced AI models. Powered by UniTS, a state-of-the-art (SoTA) universal time-series model, our framework autonomously learns each patient’s unique physiological and behavioral patterns, detecting subtle deviations that signal potential health risks. Unlike classification methods that require impractical, continuous labeling in real-world scenarios, our approach uses anomaly detection to provide real-time, personalized alerts for reactive home-care interventions. 
Our approach outperforms 12 SoTA anomaly detection methods, demonstrating robustness across both high-fidelity medical devices (ECG) and consumer wearables, with a $\sim22\%$ improvement in F1 score. However, the true impact of AI on the Pulse lies in @HOME, where it has been successfully deployed for continuous, real-world patient monitoring. 
By operating with non-invasive, lightweight devices like smartwatches, our system proves that high-quality health monitoring is possible without clinical-grade equipment. 
Beyond detection, we enhance interpretability by integrating LLMs, translating anomaly scores into clinically meaningful insights for healthcare professionals.
\end{abstract}


%%
%% The code below is generated by the tool at http://dl.acm.org/ccs.cfm
%%


\maketitle
\input{sections/introduction}

\input{sections/related_work}

\input{sections/method}

\input{sections/results}

\input{sections/conclusion}

% \begin{acks}
% We thank all patients and collaborators involved in @HOME. This work was partially supported by [Funding Agency-Grant/Award Number].
% \end{acks}

\section{GenAI Usage Disclosure}
In accordance with the ACM policy, we disclose limited use of generative AI tools during manuscript preparation. Specifically, ChatGPT was employed for light editorial feedback, such as refining grammar, improving sentence clarity, and suggesting alternative phrasings. All substantive content, including technical descriptions, analyses, and conclusions, was authored entirely by the researchers.

\bibliographystyle{ACM-Reference-Format}
\bibliography{mybib}

\end{document}

%% file: sections/introduction.tex
\section{Introduction}
Continuous patient monitoring is crucial for early detection of health deterioration, risk identification, and clinical decision support~\cite{taskasaplidis2024stress}. Chronic conditions such as Alzheimer's and Parkinson's disease lead to progressive declines in cognitive, motor, and autonomic functions, disrupting daily life and well-being~\cite{Heilman2022EmotionalAN}. Traditional healthcare monitoring relies on episodic assessments, leaving fluctuations in vital signs (e.g., heart rate, respiratory rate, sleep patterns), daily activities (e.g., movement patterns, bathroom usage), and environmental conditions (e.g., CO$_2$ levels, humidity, luminosity) largely untracked~\cite{parvin2019personalized}. Stress further exacerbates disease progression, driven by physiological, psychological, and environmental stressors such as mood changes, social isolation, and uncertainty about disease outcomes~\cite{kraut2022mental}. Real-time monitoring of stress and health anomalies is essential for symptom management and personalized interventions.

We present an anomaly detection system that uses UniTS~\cite{gao2025units}, a universal time-series model, to learn patient-specific baselines from wearable and ambient data. Unlike classification models, our approach detects subtle deviations in real-time without predefined labels. The initial UniTS model does not incorporate contextual features for anomaly detection. We enhance and adapt UniTS to enable \emph{contextual anomaly detection} by incorporating auxiliary inputs and applying selective masking techniques. Currently deployed in a clinical study, the system continuously monitors patients in home-care environments, providing real-time alerts to healthcare professionals.

\noindent\textbf{Challenges in Real-World Patient Monitoring.}
Long-term monitoring presents challenges across multiple domains. Variations in vital signs (HR, HRV, respiration, sleep cycles) may indicate stress, cardiovascular risks, or autonomic dysfunction. Behavioral deviations, such as irregular sleep patterns or unusual bathroom usage, can signal disease progression, dehydration, or infections. Environmental factors, including humidity and CO$_2$ fluctuations, contribute to sleep disturbances, respiratory stress, and cognitive fatigue~\cite{prenkaj2023unsupervised}. Ensuring seamless, non-invasive monitoring remains a key hurdle. While wearable and ambient sensors provide passive real-time data collection, device limitations and environmental noise impact signal reliability~\cite{taskasaplidis2024stress}. Moreover, effective clinical integration requires AI-generated anomaly detection alerts to be interpretable; without clear explanations, AI-driven insights may be misinterpreted or ignored~\cite{xie2024human}.

\noindent\textbf{Proposed Approach: A Multi-Signal Anomaly Detection System for Patient Monitoring.}
Our system integrates physiological and environmental data to provide a comprehensive patient monitoring framework, identifying deviations from each patient's \textit{individual baseline}.
Additionally, to enhance real-time clinical decision-making, our system automatically reports detected anomalies to healthcare professionals through an intelligent alert system. Integrated LLMs generate human-readable explanations, tailoring responses to match each clinician's expertise, terminology, and workflow. We argue that this personalization improves the usability and trustworthiness of AI-driven insights.
Designed for deployment in both clinical and home settings, our system is currently monitoring in a pilot study 6 patients with neuro-degenerative conditions across diverse demographics\footnote{Code is available at \href{https://github.com/davegabe/ai-on-the-pulse}{https://github.com/davegabe/ai-on-the-pulse}}.

%% file: sections/related_work.tex
\section{Related Work}
Our study contributes to \emph{anomaly detection in multivariate time series for patient monitoring}. While much research has focused on emotion and stress detection, fewer works leverage unsupervised anomaly detection of vital signs, essential for real-world healthcare applications.

\noindent\textbf{Anomaly Detection in Multivariate Time Series.} Detecting anomalies in multivariate time series is a core challenge in machine learning, with applications in healthcare, finance, and industrial monitoring. Traditional statistical and distance-based methods struggle with high-dimensional data, leading to the adoption of deep learning approaches. Reconstruction-based models, such as DAGMM \cite{zong2018deep}, OmniAnomaly \cite{su2019omni}, MSCRED \cite{zhang2019deep}, and  USAD \cite{audibert2020usad} use autoencoders to detect deviations. Adversarial learning, explored in MAD-GAN \cite{li2019madgan}, TadGAN \cite{geiger2020tadgan} and HypAD \cite{flaborea2023hypad}, refines anomaly detection through synthetic sequences but requires careful training for stability. Graph-based models, such as MTAD-GAT \cite{zhao2020multivariate} and GDN \cite{deng2021graph}, leverage attention mechanisms to model dependencies between variables.

\noindent\textbf{Anomaly Detection in Patient Monitoring.} 
The impracticality of acquiring consistently labeled data for patient health deterioration makes unsupervised anomaly detection a practical solution, identifying irregularities that diverge from normal patterns. Methods utilizing autoencoders on physiological data from MIoT devices detect and classify abnormalities \cite{abououf2023explainable}. Other work applies the Contextual Matrix Profile to activity data to identify anomalies indicative of adverse health conditions \cite{bijlani2022unsupervised}.

\noindent\textbf{Explainability in Anomaly Detection.} Explainability in AI-driven anomaly detection is essential for real-world adoption, particularly in healthcare applications where trust and transparency are paramount. While many anomaly detection models rely solely on reconstruction errors or classification probabilities, recent work has integrated explainability frameworks to provide more interpretable results. Graph-based models, such as GDN \cite{deng2021graph}, offer some level of interpretability by visualizing inter-variable relationships. Similarly, methods like CAE-M \cite{zhang2021caem} incorporate memory-augmented networks to retain prototypical normal patterns, making it easier to compare deviations. However, these approaches still fall short in providing human-readable explanations. A more recent approach involves integrating LLMs for explainability. The use of LLMs allows for the translation of anomaly scores into domain-specific explanations tailored for clinicians  \cite{xie2024human}. This aligns with our approach, where we fine-tune UniTS \cite{gao2025units} for anomaly detection and integrate LLM-generated natural language explanations to enhance interpretability.

%% file: sections/method.tex
\section{Methodology}\label{sec:method}
\subsection{Problem Definition}\label{sub:problemdefinition}
Anomaly detection in multivariate time series \cite{flaborea2023hypad,prenkaj2023unsupervised} is the process of identifying observations within a sequence of multivariate time-dependent data that deviate significantly from the individually learned patterns of normal behavior. This is important in the context of monitoring individuals with neuro-degenerative diseases, where deviations in physiological signals can indicate changes in health status.

\noindent\textbf{Multivariate Anomaly Detection.} Let $\mathbf{X} = {\mathbf{x}_1, \dots, \mathbf{x}_T}$ be a multivariate time series with $\mathbf{x}_t \in \mathbb{R}^n$ as the $n$-dimensional feature vector at time $t$. The model $f$, trained on $\mathbf{X}_{\text{train}} = {\mathbf{x}_1, \dots, \mathbf{x}_t}$ ($t < T$), learns normal behavior and assigns an anomaly score $S_t = g(\mathbf{x}_t, f(\mathbf{X}_{\text{train}}))$ where $g(\cdot)$ 
%is a function that 
quantifies the deviation between $\mathbf{x}_t$ and the learned normal pattern. A data point $\mathbf{x}_t$ is classified as anomalous if its anomaly score exceeds a (patient-dependent) threshold $\tau$.

\noindent\textbf{Temporal Modeling.} To model dependencies, we use a sliding window $W_t = {\mathbf{x}_{t-K+1}, \dots, \mathbf{x}_t}$ of length $K$. The full sequence is represented as overlapping windows $\mathbf{W} = {W_1, \dots, W_T}$. Let $W$ denote a training window and $\widehat{W}$ an unseen test window.

\noindent\textbf{Binary Classification of Anomalies.} Given an unseen window $\widehat{W}_t$ for $t > T$, the anomaly detection model assigns a binary label $y_t = 1$ indicating whether an anomaly has been detected if $S_t > \tau$; otherwise $y_t = 0$.
\begin{figure}[!t]
    \centering
    \includegraphics[width=\linewidth]{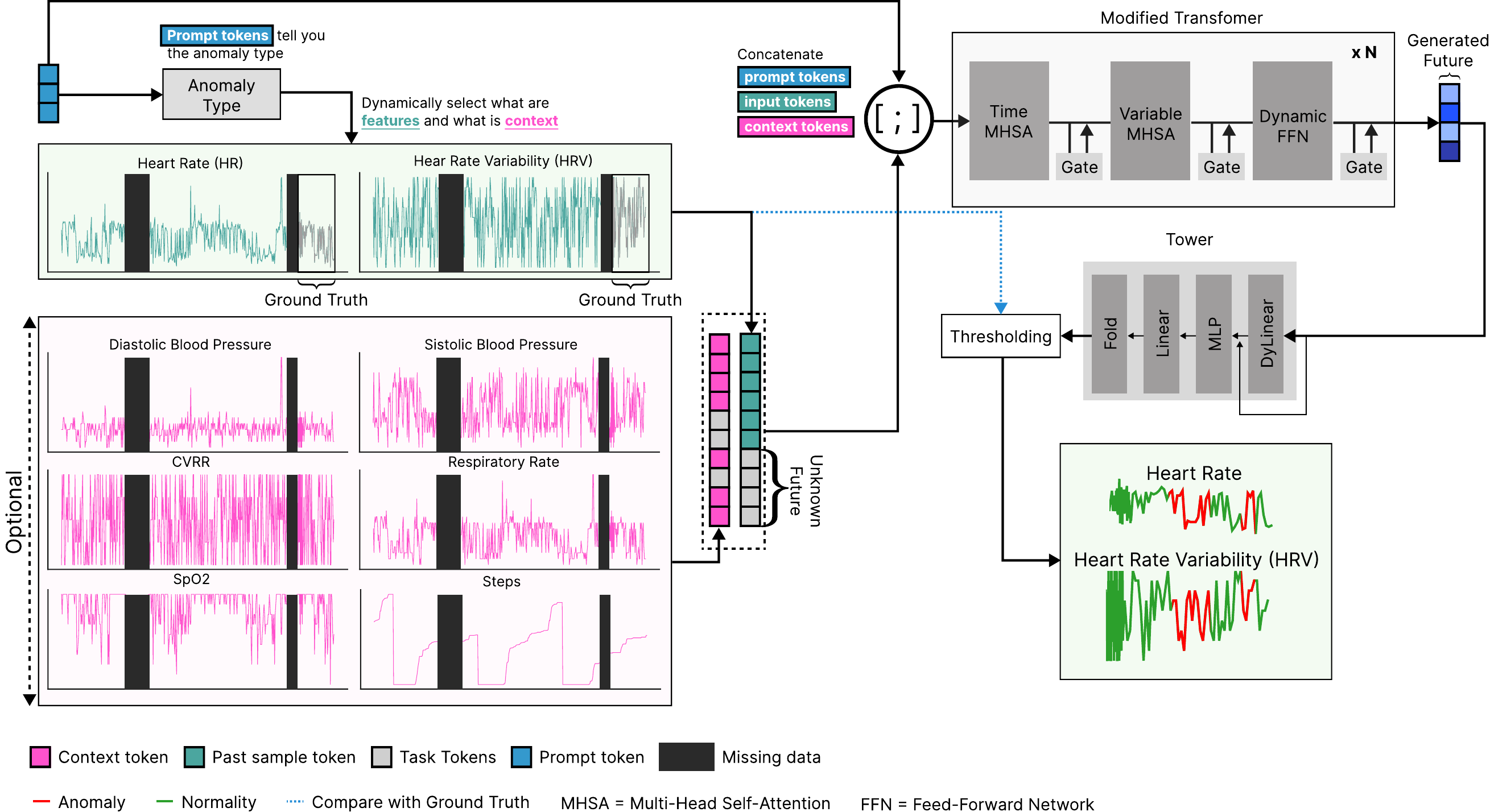}
    \caption{\noindent\textbf{Fine-tuning UniTS for Contextual Anomaly Detection.} Inputs are tokenized into past, future, and contextual segments, with missing values masked via task tokens. Based on the anomaly type, relevant input, context, and prompt tokens are selected, concatenated, and passed through $N$ transformer blocks. Future predictions are reconstructed and compared to ground truth; anomalies are flagged using a threshold.}
    \label{fig:units}
\end{figure}

\subsection{Framework Overview}\label{sub:framework}

UniTS~\cite{gao2025units} is a foundation model for multivariate time series, supporting forecasting, imputation, classification, and anomaly detection. It uses self-attention to capture dependencies across time and features, making it effective for multi-sensor data. However, the original UniTS formulation lacks the notion of contextual features in anomaly detection task. We extend and fine-tune UniTS to support \emph{contextual anomaly detection} by integrating auxiliary signals and selective masking.

\noindent\textbf{Fine-tuning UniTS for Anomaly Detection.}
We first fine-tuned UniTS on wearable-derived heart rate (HR) and heart rate variability (HRV) to detect stress-induced anomalies.  To better emulate the behavior of noninvasive wearable devices, we preprocessed the data set by reducing the sampling rate to reflect the practical constraints of smartwatch data acquisition. The model was fine-tuned for 20 epochs with an exponential learning rate decay. In real-world settings, such as in \textit{@HOME}, richer inputs --such as respiration, physical activity, and cardiovascular metrics can further help distinguish true anomalies from expected physiological changes.

\noindent\textbf{Adapting UniTS for Contextual Anomaly Detection.}
We reformulate anomaly detection as an imputation task: future values of target features in $\mathbf{X}$ are masked, while past values and contextual tokens are preserved. This setup allows the model to learn dependencies between auxiliary signals and the targets, e.g., linking elevated HR to physical activity. Prompt tokens condition the model on the anomaly type. Training uses a reconstruction loss; at inference, the reconstruction error on masked target features serves as the anomaly score.

\noindent\textbf{Input Tokenization.}
Inputs are tokenized into \textit{sample}, \textit{prompt}, and \textit{task} tokens. Given $\mathbf{X} \in \mathbb{R}^{T \times n}$, we split into past ($\mathbf{X}_P$) and future ($\mathbf{X}_F$) segments:
\[
z_{\mathbf{X}_P} \in \mathbb{R}^{\frac{T - \lambda}{k} \times n \times d}, \quad z_{\mathbf{X}_F} \in \mathbb{R}^{\frac{\lambda}{k} \times n \times d}.
\]
Context $\mathbf{C} \in \mathbb{R}^{T \times m}$ is tokenized as:
\[
z_\mathbf{C} \in \mathbb{R}^{\frac{T}{k} \times m \times d}.
\]
Tokens are concatenated along time ($\mathcal{T}$) and feature ($\mathcal{F}$) axes:
\[
z_{\mathbf{X},\mathbf{C}} = [[z_{\mathbf{X}_P};z_{\mathbf{X}_F}]_\mathcal{T};z_\mathbf{C}]_\mathcal{F} \in \mathbb{R}^{\frac{T}{k} \times (n+m) \times d}.
\]

\noindent\textbf{Prompt and Task Tokens.}
Prompt tokens $z_p \in \mathbb{R}^{p \times n \times d}$ encode task-specific goals (e.g., “detect tachycardia”) and are prepended:
\[
\mathbf{Z} = [z_p; z_{\mathbf{X},\mathbf{C}}]_\mathcal{T} \in \mathbb{R}^{(p + \frac{T}{k}) \times (n+m) \times d}.
\]
\emph{GEN tokens}, originally used for imputation, are repurposed to reconstruct masked future values and fill missing data, enabling contextual anomaly scoring.

\noindent\textbf{Model Components.}
The input $\mathbf{Z}$ is processed through $N$ transformer blocks, each tailored for multivariate time series. A dual self-attention mechanism, \emph{Multi-Head Self-Attention (MHSA)}, captures dependencies across both time and variable dimensions. A \emph{Dynamic Feed-Forward Network (FFN)} then models token-wise interactions more effectively than standard FFNs. \emph{Gating modules} re-scale latent features dynamically, improving stability and reducing interference. Finally, the \emph{Tower} module reconstructs the future segment $\hat{\mathbf{X}}F$ from predicted tokens $\hat{z}{\mathbf{X}_F}$ using:
  \[
  \hat{\mathbf{X}}_F = \text{Proj}(\text{MLP}(\hat{z}_{\mathbf{X}_F} + \text{DyLinear}(\hat{z}_{\mathbf{X}_F}))).
  \]
where $\text{DyLinear}(z_t, w) = \text{Interp}(w) \cdot z_t$ applies interpolation-based smoothing. Together, these components enable UniTS to learn rich, context-aware representations for robust anomaly detection.

\noindent\textbf{Medical Interface.} \textit{@HOME} has been deployed within our clinical partner’s patient management system, showing anomaly scores detected by the model with visualizations and LLM-generated explanations based on medical records and anomaly data.

%% file: sections/results.tex
\section{Datasets and Experiments}

\subsection{Stress detection datasets: PGP Sensors vs. Invasive ECG}\label{sec:wesad}

In our initial experiments, our primary objective was to determine wether data obtained from non-invasive technology -- such as those using PPG sensors -- could effectively replace ECG-derived data for detecting abnormal physiological states, such as stress. We focused on HR and HRV features and evaluated the capability of SoTA models to detect stress anomalies. We benchmarked ECG-based models using DREAMER~\cite{katsigiannis2017dreamer} and HCI~\cite{soleymani2011multimodal}, and used WESAD~\cite{schmidt2018introducing} -- containing both ECG and BVP -- to directly compare invasive and non-invasive signals (see~\cref{tab:comparison}).
Following~\cite{10.1145/2674396.2674446}, we downsample signals to extract HR and HRV every 10 seconds using a 60-second sliding window to emulate smartwatch-like conditions.

\subsection{@HOME: In-House Patient Monitoring Dataset}\label{sec:home}
\begin{figure}
\centering
\begin{subfigure}{\linewidth}
    \centering
    \includegraphics[width=\linewidth]{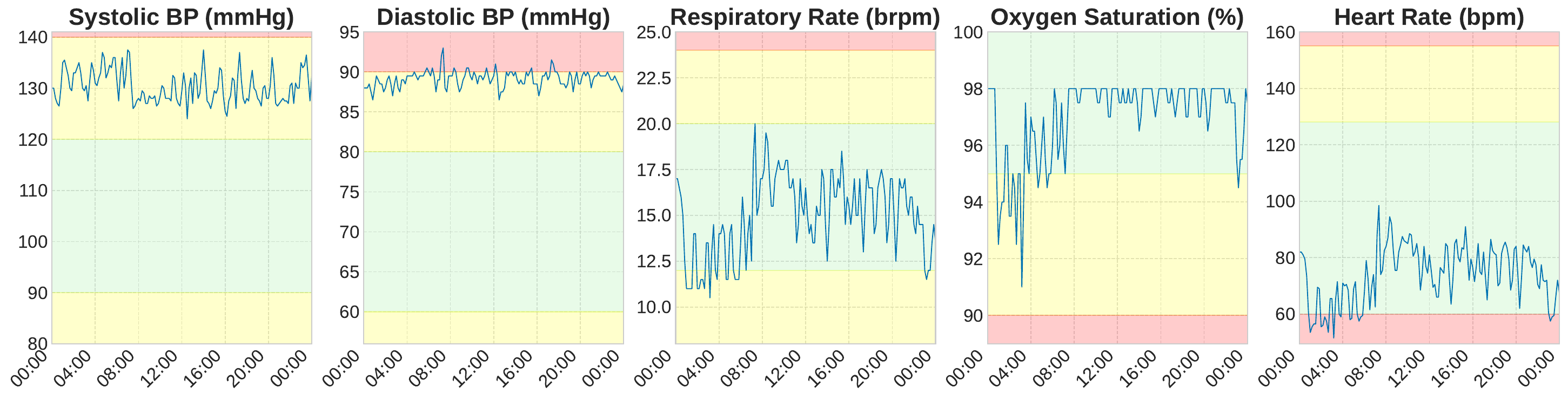}
\end{subfigure}
\begin{subfigure}{\linewidth}
    \centering
  \includegraphics[width=\linewidth]{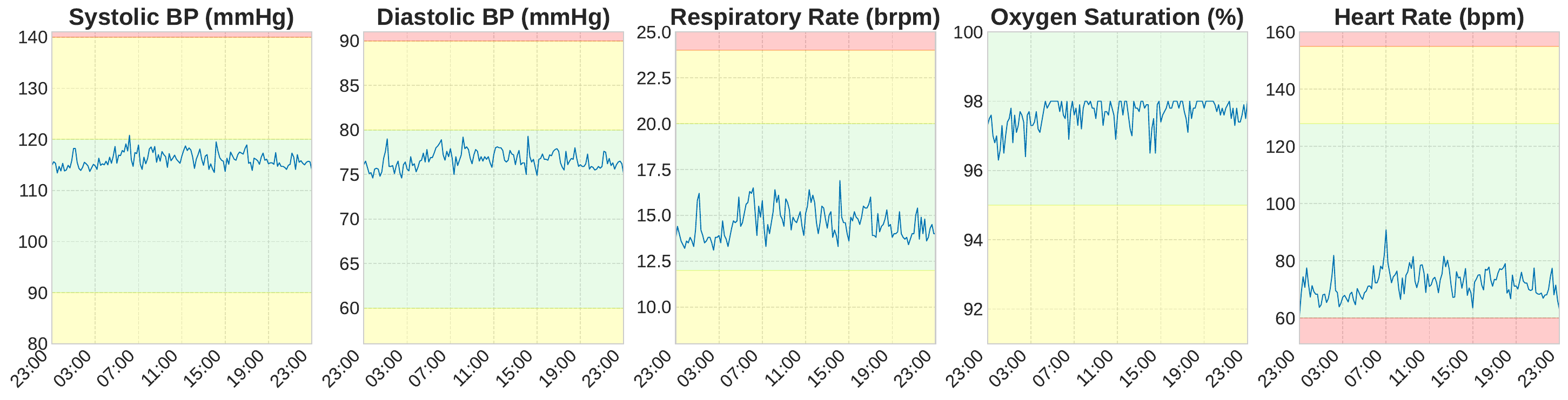}
  \label{fig:sub1}
\end{subfigure}
    \caption{@HOME data samples: pre-pilot (top) and pilot (bottom). Shaded regions -- green (normal), yellow (concern), red (abnormal) -- reflect  severity levels for each vital sign. \label{fig:dataset}}
\end{figure}

Currently, naturalistic datasets collecting a wide range of vital and environmental signs are not available. We fully evaluate our approach on \textit{@HOME}, a real-world dataset collected via consumer-grade wearables in a long-term home monitoring setup. Devices sampled data every minute and were chosen for their usability (5-day battery, fast charging). The recorded signals include HR, HRV, CVRR, Respiration Rate, SpO2, BP (systolic/diastolic), Steps, Sleep Phases, and a rich set of environmental metrics: Temperature, Humidity, CO$_2$, Air Pressure, Luminosity, TVOC, and Patient Room Location (bathroom, kitchen, living room, study, bedroom).
The study tracked in their homes 6 elderly patients with early-stage neurological conditions at 1-minute resolution. 
As an example, ~\cref{fig:dataset} shows data from two patients -- one from the pre-pilot and one from the pilot -- highlighting how individual baselines can vary significantly. For instance, the pre-pilot patient's systolic/diastolic blood pressure is consistently flagged as slightly or highly abnormal (yellow/red), despite being stable over time and representing their personal normality. Our model instead adapts to \textit{individual baselines}, improving detection of true anomalies.

\noindent\textbf{Missing signals and interpolation.}  
Long-term patient monitoring often suffers from missing data due to sensor issues or connectivity loss, distorting time-dependent signals. We evaluate interpolation methods using MASE for multivariate accuracy and DTW for temporal alignment. Among the tested approaches, simple methods like nearest-neighbor and nearest-window %perform best, showing the lowest MASE (\(1.1 \times 10^{-2}\)) and DTW (\(3.74 \times 10^{1}\)) scores, while
outperform more complex methods like LOESS and spline interpolation. We choose nearest-window for its strong performance and better shape preservation compared to standard nearest-neighbor. We use 16-sample input windows and limit interpolation to cases with $\leq 5$ missing samples to minimize distortion.

\begin{table}[!t]
\centering
\resizebox{\linewidth}{!}{%
    \begin{tabular}{@{}lcccc|c@{}}
    \toprule
     
    {} &
      {DREAMER~\cite{katsigiannis2017dreamer}} &
      {HCI~\cite{soleymani2011multimodal}} &
      {\begin{tabular}[c]{@{}c@{}}WESAD~\cite{schmidt2018introducing}\\ (ECG)\end{tabular}} &
      {\begin{tabular}[c]{@{}c@{}}WESAD~\cite{schmidt2018introducing}\\ (BVP)\end{tabular}} &
      {\begin{tabular}[c]{@{}c@{}}AVG\\ F1\end{tabular}} \\ \midrule
    LSTM-NDT~\cite{hundman2018detecting}    & 0.235 & 0.373 & 0.694 & 0.760 & 0.515 ± 0.218 \\
    MSCRED~\cite{zhang2019deep}      & 0.434 & 0.618 & 0.594 & 0.536 & 0.545 ± 0.071 \\
    HypAD~\cite{flaborea2023we}       & 0.513 & 0.547 & 0.669 & 0.479 & 0.552 ± 0.072 \\
    TranAD~\cite{tuli2022tranad}      & 0.515 & 0.622 & 0.474 & 0.717 & 0.582 ± 0.095 \\
    OmniAnomaly~\cite{su2019robust} & 0.543 & 0.513 & 0.716 & 0.583 & 0.589 ± 0.078 \\
    MTAD-GAT~\cite{zhao2020multivariate}    & 0.369 & 0.677 & 0.757 & 0.676 & 0.620 ± 0.149 \\
    CAE-M~\cite{zhang2021unsupervised}      & 0.634 & 0.513 & \underline{0.762} & 0.590 & 0.625 ± 0.090 \\
    MAD-GAN~\cite{Li2019MADGANMA}     & 0.390 & \underline{0.678} & 0.735 & 0.750 & 0.638 ± 0.146 \\
    TadGAN~\cite{geiger2020tadgan}      & 0.528 & 0.657 & 0.705 & 0.680 & 0.643 ± 0.068 \\
    USAD~\cite{audibert2020usad}        & 0.590 & 0.641 & 0.686 & 0.759 & 0.669 ± 0.062 \\
    GDN~\cite{deng2021graph}         & \underline{0.712} & 0.501 & 0.697 & \underline{0.778} & 0.672 ± 0.103 \\
    DAGMM~\cite{zong2018deep}       & 0.619 & 0.596 & 0.744 & 0.733 & \underline{0.673 ± 0.066} \\
    \midrule
    UniTS~\cite{gao2025units}       & \textbf{0.828} & \textbf{0.864} & \textbf{0.793} & \textbf{0.800} & \textbf{0.821 ± 0.049} \\
    \bottomrule
    \end{tabular}%
}
\caption{F1 scores for stress detection using HR and HRV features across four datasets and SoTA models. Remarkably, our solution achieves the highest average performance across all datasets and smaller std dev. Bold values indicate the best results, underlined values indicate the second best.
\label{tab:comparison}}
\end{table}

\subsection{Experimental Setup}
\noindent\textbf{Anomaly Detection Model Experiment.} 
We compare UniTS against 12 SoTA methods (see~\cref{tab:comparison}) using their default hyperparameters. UniTS was fine-tuned with window size 5, embedding dimension 128, and learning rate $5\times10^{-4}$. We split each series 80:20 for train/test, trained for 20 epochs, and report the F1 score averaged over 5-fold CV. Anomalies are defined as points below the 3rd or above the 97th percentile. For WESAD, the only detectable anomaly was \textit{stress} since no contextual data was provided.

\noindent\textbf{Contextual Anomaly Detection Experiment.}
On @HOME, models were fine-tuned \textit{for each patient}, for 2 epochs with windows of 16 samples, where the first 8 samples represent past tokens and the latter 8 are masked to be imputed. Thresholds for detection are set adaptively using a POT-based method~\cite{siffer2017anomaly}, enabling robust identification of high-error events. 
Given the inherent difficulty of obtaining definitive ground truth labels for anomalies, our evaluation protocol mirrors a production environment: detected anomalies are forwarded to a medical expert, and explanations generated via GPT-4o are provided alongside each anomaly for review. This procedure allows us to monitor both true positive and false positive rates, ensuring that the system not only identifies anomalies, but also provides clinically meaningful insights.

\subsection{Results}

\noindent\textbf{UniTS surpasses SoTA anomaly detection systems and shows remarkable stability and robustness across datasets.} As shown in~\cref{tab:comparison}, UniTS consistently outperforms all compared systems, achieving a notable average improvement of $22\%$ over the second-best -- $16.29\%$ on DREAMER, $27.43\%$ on HCI, $4.07\%$ on WESAD (ECG), and $2.83\%$ on WESAD (BVP). BVP signals are prone to noise and artifacts~\cite{10.1145/3027063.3053121}, but UniTS handles this via attention over time and features, sustaining high performance.\footnote{UniTS yields an FPR of $0.031$ across both WESAD versions, and FNRs of $0.172$ (ECG) and $0.159$ (BVP), indicating high recall.}

\noindent\textbf{UniTS provides reliable anomaly detection with strong clinical validation.} In the \textit{@HOME} deployment, anomaly reports -- comprising visualized signals and AI-generated explanations\footnote{The LLM is guided to interpret input signals, providing reasoning for abnormal readings and expected patterns using clinically appropriate medical language. Examples of the prompt and generated explanations are provided at the \textit{@HOME} link \href{https://github.com/davegabe/ai-on-the-pulse}{https://github.com/davegabe/ai-on-the-pulse} .} -- were assessed by a highly experienced senior geriatrician, who continuously monitors the experiment and the selected patients, ensuring expert and consistent evaluation despite being the sole assessor. Only true and false positives were considered, as false negatives require exhaustive review. Results from the pre-pilot and the pilot, despite data noise, show that 93.75\% of 32 anomalies\footnote{Only 32 anomalies have been detected during the first 3 months, which is a reasonable number since anomalies, by definition, are rare events.} (\cref{tab:anomaly_summary}) were confirmed true positives; the remaining 6.25\% were attributed to sensor issues. The medical expert consistently rated the detected anomalies as clinically meaningful (significance $\geq$ 3.0), but moderate criticality scores ($\sim$2.0–2.4) across all types indicate that no severe events occurred during the monitoring period.

\begin{table}
    \centering
    \resizebox{\linewidth}{!}{
    \begin{tabular}{@{}lcccc@{}}
    \toprule
    \textbf{Anomaly Type} & \textbf{Quantity} & \textbf{Sensor Error} & \textbf{Significance} & \textbf{Criticality} \\ \midrule
    Hyper/Hypotension     & 10 & 0           & 3.00 $\pm$ 0.63       & 2.30 $\pm$ 0.64      \\
    Abn. HRV              & 9  & 0           & 3.22 $\pm$ 0.42       & 2.33 $\pm$ 0.47      \\
    Stress                & 10 & 0          & 3.20 $\pm$ 0.40        & 2.40 $\pm$ 0.49      \\
    Sleep Quality         & 3 & 2          & 3.00 $\pm$ 0.00       & 2.00 $\pm$ 0.00      \\ \bottomrule
    \end{tabular}
    }
    \caption{Assessment of ratings given by a medical expert for anomalies detected by our system. Note also that some anomaly types have never been reported in deployment. The scores are in $[0,5]$.\label{tab:anomaly_summary}}
\end{table}

%% file: sections/conclusion.tex
\section{Conclusion and Implications}

This work introduces \emph{AI on the Pulse}, a real-world-ready anomaly detection system designed for continuous patient monitoring in clinical and home-care settings. 
By combining wearable sensors, ambient intelligence, and the UniTS model, it learns individual patient baselines to detect subtle health changes early. Beyond outperforming SoTA methods, its practical value is shown in @HOME, where it provides real-time, interpretable alerts using consumer-grade devices, removing the need for clinical-grade equipment. Designed with clinicians in mind, its explainability fosters trust and usability. Future work includes broader deployment, more sensor integration, and scaling to full clinical use.